# Sky-GVINS: a Sky-segmentation Aided GNSS-Visual-Inertial System for Robust Navigation in Urban Canyons

Jie Yin[a], Tao Li[a], Hao Yin[a], Wenxian Yu[a] and Danping Zou[a]

[a]Shanghai Key Laboratory of Navigation and Location Based Services, Shanghai Jiao Tong University, Shanghai, China

ABSTRACT

Integrating Global Navigation Satellite Systems (GNSS) in Simultaneous Localization and Mapping (SLAM) systems draws increasing attention to a global and continuous localization solution. Nonetheless, in dense urban environments, GNSS-based SLAM systems will suffer from the Non-Line-Of-Sight (NLOS) measurements, which might lead to a sharp deterioration in localization results. In this paper, we propose to detect the sky area from the up-looking camera to improve GNSS measurement reliability for more accurate position estimation. We present Sky-GVINS: a sky-aware GNSS-Visual-Inertial system based on a recent work called GVINS. Specifically, we adopt a global threshold method to segment the sky regions and non-sky regions in the fish-eye sky-pointing image and then project satellites to the image using the geometric relationship between satellites and the camera. After that, we reject satellites in non-sky regions to eliminate NLOS signals. We investigated various segmentation algorithms for sky detection and found that the Otsu algorithm reported the highest classification rate and computational efficiency, despite the algorithm's simplicity and ease of implementation. To evaluate the effectiveness of Sky-GVINS, we built a ground robot and conducted extensive real-world experiments on campus. Experimental results show that our method improves localization accuracy in both open areas and dense urban environments compared to the baseline method. Finally, we also conduct a detailed analysis and point out possible further directions for future research. For detailed information, visit our project website at https://github.com/SJTU-ViSYS/Sky-GVINS.

CONTACT    Danping Zou    ✉ dpzou@sjtu.edu.cn

# 1. Introduction

Intelligent navigation is an essential function in a wide range of applications including autonomous driving and robot navigation. Simultaneous Localization and Mapping (SLAM) is the most widely used algorithm in these applications to track the location while constructing a map of the environment simultaneously. As indicated in some works (Aulinas et al. 2008; Cadena et al. 2016), a lot of SLAM systems have been proposed with diverse sensor settings. Although LiDAR-based SLAM is usually considered for its superior performance, LiDAR is expensive and sometimes not suitable for lightweight devices. Therefore, there is increasing interest in using low-cost and lightweight sensors like cameras and Inertial Measurement Unit (IMU) in practice.

Generally speaking, camera-based or visual SLAM can be grouped into monocular and stereo according to the number of cameras. Monocular SLAM suffers from scale uncertainty and scale drift. Stereo cameras take advantage of the known baseline distance to calculate metric depth by triangulation. It is still hard for a pure vision-based SLAM to work well in complicated scenes with poor texture, motion blur, and occlusions. Therefore, visual-inertial algorithms have been proposed to enhance the robustness in these complicated environments. Some visual-inertial systems like ORB-SLAM3 (Campos et al. 2021) and VINS-Mono (Qin et al. 2018) perform well in public SLAM datasets including EUROC (Burri et al. 2016) and TUM VI (Schubert et al. 2018), their performances drop sharply in large-scale outdoor environments as discussed in our previous work (Yin et al. 2021).

To improve localization accuracy in these scenarios, one solution is to couple Global Navigation Satellite Systems (GNSS) to SLAM systems for a drift-free and global-aware positioning result (Cao et al. 2021; Li et al. 2020). However, the performance of GNSS-based systems highly depends on satellite visibility. Most researches focus on mitigating multi-path effects, which can be detected by the elevation angle or Signal to Noise Ratio (SNR). Actually, in urban canyons, NLOS (Non-Line-Of-Sight) signals occur frequently – i.e., signals received with no direct ray but from a reflection path. NLOS signals will degrade localization accuracy because the propagation time delays will produce an additional error on the pseudo-range estimation. This error is typically tens of meters or even greater. There are several NLOS mitigation methods as discussed in (Groves et al. 2013). Nonetheless, they have their limitations like dependency on the previously known 3D city model or high-cost antennas.

To address the aforementioned issues, we present Sky-GVINS: a sky-aware GNSS-Visual-Inertial system with NLOS mitigation. We highlight the major contributions as follows: Firstly, we introduce a lightweight method that uses the global threshold algorithm on the sky-pointing images to separate LOS and NLOS signals. Our approach is efficient and effective, and can be easily applied to existing GNSS-based systems. Furthermore, we present a robust SLAM system that can obtain accurate global state estimation both in dense urban environments and open-sky regions, which outperforms other existing systems in real-world experiments. The novelty lies in implementation details such as estimating the relative transformation between the satellite and the camera on the fly and NLOS satellite detection via sky segmentation.

The remaining paper is organized as follows: Section 2 reviews multi-sensor SLAM, introduces a few approaches to mitigate the NLOS effect, and lists some applications of sky segmentation; Section 3 explains implementing details about our proposed system; After that, Section 4 describes the experiments, results and analysis; Finally, we summarize our contributions and prospect our future research directions in Section 5.

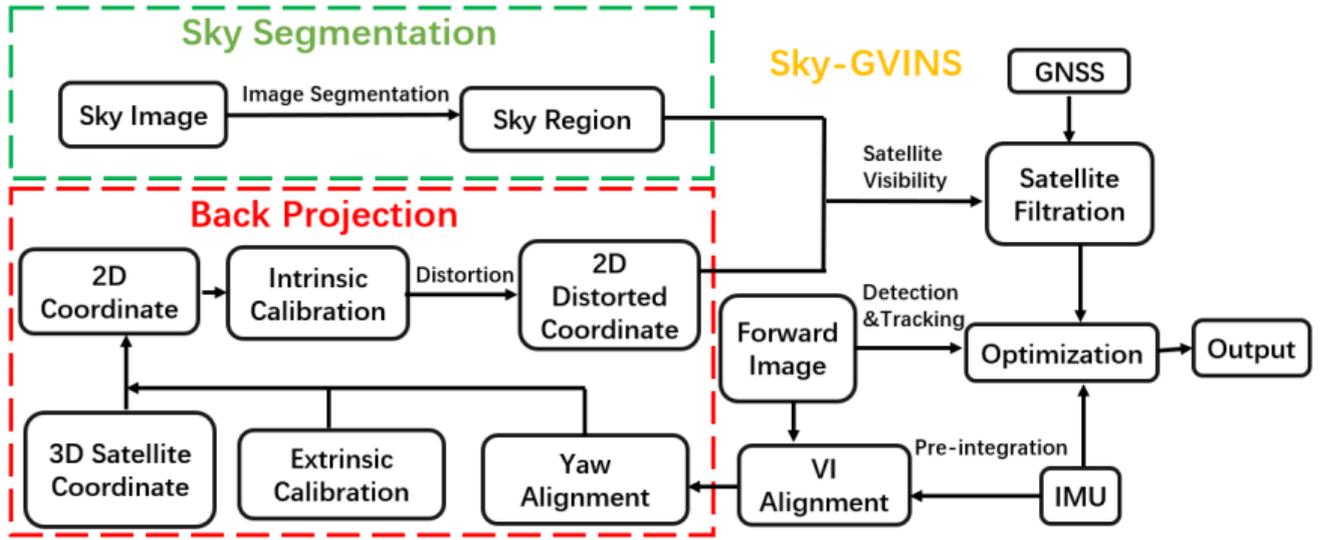

Figure 1. The diagram shows the pipeline of Sky-GVINS. The main contributions highlight the sky segmentation and back-projection part

## 2. Related Work

### 2.1. Multi-sensor SLAM

Multi-sensor fusion has been proven practical and effective in enhancing the accuracy and robustness of SLAM systems (Liu et al. 2019a, 2021b). Extensive related works have been done so far. For example, StructVIO (Zou et al. 2019) and VINS-Mono (Qin et al. 2018) integrate vision and IMU to eliminate scale uncertainty.

More recently, SLAM systems tightly coupling GNSS are attracting attention for their unique advantages of high accuracy and robustness in outdoor scenarios. For example, Li' work (2021) tightly couples GNSS and LiDAR for precise vehicle navigation. (Liu et al. 2021) tightly couples vision, IMU and GNSS and outperforms state-of-the-art visual-inertial systems in complex urban scenes. The latest relevant work is GVINS (Cao et al. 2021), a GNSS, vision and IMU tightly-coupled system, which can gain from even a single satellite while conventional GNSS-based methods need at least four. However, these GNSS-based systems will suffer from the NLOS measurements and obtain a worse localization accuracy in dense urban environments. To address this issue, it could be better to eliminate NLOS signals before sensor fusion.

### 2.2 Elimination of NLOS Signals

GNSS positioning in urban canyons remains a challenge because surrounding buildings and trees might block, reflect and diffract satellite signals. Researchers have proposed many approaches to address those challenges, among which NLOS mitigation has been a popular one. There are two main approaches to classifying NLOS/LOS signals. The first is GNSS self-maintained approaches, including dual-polarized antennas (Jiang et al. 2014), RAIM (Receiver Autonomous Integrity Monitoring) algorithm (Jiang et al. 2011), and machine learning methods like decision tree or support vector machine (Obst et al. 2012). Nonetheless, the RAIM algorithm takes effect only when the majority of the received signals are LOS signals, while the learning-based approach is only suitable for specific scenarios that are similar to the pre-training environment. The second is environmental feature-aided approaches, such as the 3D Map-Aided (3DMA) method (Obst et al. 2012). More simply, (Marais et al. 2014) proposed a method based on image processing and detection of the sky and non-sky areas, which reports a promising accuracy. Similarly, our system adopts an image-based method for NLOS mitigation for simplicity and efficiency. However, the transformation between the camera and GNSS receivers is a critical parameter that can not be determined offline. Different from some works which use dual antennas for orientation (Brazeal et al. 2021), our method tightly couples VIO for estimating the transformation relationship of the camera and the receiver, which will be described in Section 3.3. Meanwhile, our system can output high-accuracy global trajectory with the aid of VIO even when the sensor platform enters a GNSS-denied environment for a while.

*2.3 Sky Segmentation*

Segmentation of a digital image is the process of its division into several disjoint regions so that pixels of every area have similar visual characteristics. It is a practical approach to simplify the representation of an image for further analysis. Commonly used conventional image segmentation algorithms include threshold-based methods (Aimary et al. 2010), edge-based methods, and clustering methods. According to the uniform threshold for the whole map or different thresholds for different regions, the threshold methods can be divided into the global threshold method and the local threshold method (Ahamad et al. 1999) (also called the adaptive threshold method).

Sky segmentation is to divide the image into the sky and non-sky regions. Sky segmentation can be used for various purposes including obstacle detection and navigation. For example, McGee et al. segment the sky with a Support Vector Machine (SVM) (Burges et al. 1998) to detect obstacles for small autonomous aircraft. Gakne et al. tightly couple sky-pointing images and GNSS for navigation (Gakne et al. 2018). Thomas Stone et al. (Stone et al. 2014) explore an approach for sky segmentation using ultraviolet light K-means (Krishna et al. 1999) and Gaussian Mixture Model (GMM) (Zivkovic 2004) to produce reliable localization and orientation on the route in different weather conditions. These works inspire us to explore applying image segmentation to improve existing GNSS-SLAM systems.

## 3. Methodology

In this section, we introduce the methodology of our system. Sky-GVINS tightly couples forward-looking images, inertial information and GNSS measurements and utilizes sky-pointing images to improve GNSS measurement quality based on an image segmentation strategy. An overview of our system is illustrated in Figure 1. Firstly, the frames and state of the system are presented. Secondly, the optimization part is described. Thirdly, the initialization phase is introduced. Finally, the sky segmentation-based NLOS mitigation strategy is introduced.

*3.1. Frames and States*

Most of our the spatial frames keep the same as those in GVINS (Cao et al. 2021), so we briefly introduce them as follows:

a) Sensor Frame: Sensor frames are related to different sensors and are local frames in which sensors report their sensory data. In Sky-GVINS, sensor frames include the forward-looking camera frame $(\cdot)^f$, sky-pointing camera frame $(\cdot)^s$ and the IMU frame $(\cdot)^i$, which is chosen as the estimation target frame and denoted as body frame $(\cdot)^b$.

b) Local World Frame: Local world frame is the conventional frame in which the visual-inertial system operates as the local world frame $(\cdot)^w$. In our system, the origin of the local world frame is arbitrarily set and the z-axis is orthogonal to the ground plane.

c) ECEF Frame: The ECEF (Earth-Centered, Earth-Fixed) frame $(\cdot)^e$ is a cartesian spatial reference system. The origin

of the ECEF frame is the center of the Earth ellipsoid. The z-axis is perpendicular to the equator and orients in the north direction. The $x-y$ plane is in the equator plane with x-axis pointing to the prime meridian.

d) ENU Frame: ENU frame is a local geodetic coordinate system. The x, y, z-axis of the ENU frame $(\cdot)^n$ point to the east, north, and up directions respectively. Given a point $P$ in ECEF frame, a unique ENU frame can be determined with its origin coinciding with the point $P$.

And then we briefly review the states to be estimated in GVINS. States $\mathcal{X}$ are optimized under a sliding window and can be expressed as follows:

$$\begin{aligned} \mathcal{X} &= [x_0, x_1, \cdots x_n, \rho_0, \rho_1, \cdots \rho_m, \psi] \\ x_k &= [p_{b_k}^w, v_{b_k}^w, q_{b_k}^w, b_a, b_w, \delta t, \delta \dot{t}], k \in [0, n] \\ \delta t &= [\delta t_C, \delta t_R, \delta t_G, \delta t_E] \end{aligned} \quad (1)$$

where $p_b^w$ is the 3D position and $q_b^w$ is orientation of the body frame with respect to the local world frame. $v_b^w$ is the velocity, $b_a$ is the accelerometer bias and $b_w$ is the gyroscope bias, and $\rho$ is the inverse depth for each feature. $\psi$ is the yaw offset between the local world frame and ENU frame. $\delta t$ and $\delta \dot{t}$ are receiver clock bias and receiver clock drifting rate respectively.

### 3.2. Map Estimation

Forward images, inertial information and filtered GNSS measurements are the inputs of the optimization system. The optimum states are obtained by Maximizing A Posterior (MAP). All measurements are assumed to be independent of each other and their noise is assumed to be zero-mean Gaussian distributed, the MAP problem can be expressed as follows:

$$\begin{aligned} \mathcal{X}^* &= \arg\max_{\mathcal{X}} p(\mathcal{X} \mid z) \\ &= \arg\max_{\mathcal{X}} p(\mathcal{X}) p(z \mid \mathcal{X}) \\ &= \arg\max_{\mathcal{X}} p(\mathcal{X}) \prod_{i=0}^{n} p(z_i \mid \mathcal{X}) \\ &= \arg\min_{\mathcal{X}} \left\{ \|r_p - H_p \mathcal{X}\|^2 + \sum_{i=0}^{n} \|r(z_i, \mathcal{X})\|_{M_i}^2 \right\} \end{aligned} \quad (2)$$

where $z$ is the aggregation of $n$ measurements, $\|r_p - H_p\mathcal{X}\|^2$ represents the system state' prior information. $r(z_i, \mathcal{X})$ is the residual of each factor and $\|\cdot\|_M$ denotes the Mahalanobis norm.

Since the optimization part is not our main contribution, we suggest readers to refer Cao's work (2021) for implementing details.

*3.3. Initialization*

Firstly, the VIO initialization part is conducted using the algorithms proposed in Qin's (2018) work and thus an initial value of the scale, the gravity vector, initial IMU bias and initial velocity have been obtained. After acquiring a smooth trajectory, the three-step online GNSS-VI alignment starts as follows:

a) Coarse Anchor Point Localization: All pseudo-range measurements within the sliding window are utilized to calculate a coarse ECEF coordinate using GNSS SPP (Single Point Positioning) algorithm.

b) Yaw Offset Calibration: In the second step, the yaw offset between the ENU frame and the local world frame is calibrated using the Doppler measurement. The following optimization problem is introduced to solve the initial value of yaw offset and receiver clock rate:

$$\underset{\delta t, \psi}{\text{minimize}} \sum_{k=1}^{n} \sum_{j=1}^{p_k} \left\| r_D(\tilde{z}_{r_k}^{s_j}, \mathcal{X}) \right\|_{\sigma_{r_k,d}^{s_j}} \quad (3)$$

where $n$ is the sliding window size and $p_k$ is the satellite number observed in $k$-th epoch in the window. $r_D(\tilde{z}_{r_k}^{s_j}, \mathcal{X})$ is residual of Doppler measurement, and $\sigma_{r_k,d}^{s_j}$ is the Doppler measurement variance. Here the velocity $v_b^w$ is fixed to the VIO result and $\dot{\delta t}_k$ is assumed to be constant within the window. The coarse anchor coordinate calculated from the first step is utilized to estimate the rotation $R_n^e$ and the direction vector $\kappa_r^s$. The parameters to be estimated include the yaw offset $\psi$ and the average clock bias drift rate $\dot{\delta t}$ within the entire window. After this step, the transformation between the local world frame and ENU frame is fully calibrated.

c) Anchor Point Refinement: The previous coarse anchor point is refined and the local world trajectory with that in ECEF frame is aligned. The following problem is optimized over the sliding window measurements:

$$\underset{\delta t, p_{anc}^e}{\text{minimize}} \quad \sum_{k=1}^{n} \sum_{j=1}^{p_k} \left\| r_P(\tilde{z}_{r_k}^{s_j}, \mathcal{X}) \right\|_{\sigma_{r_k,p}^{s_j}}^2 + \sum_{k=1}^{n} \left\| r_T(\tilde{z}_{k-1}^k, \mathcal{X}) \right\|_{D_{t,k}}^2 \quad (4)$$

where $p_{anc}^e$ is the ancestor coordinate in the ECEF frame, $r_P(\tilde{z}_{r_k}^{s_j}, \mathcal{X})$ is the residual of pseudorange measurement and $r_T(\tilde{z}_{k-1}^k, \mathcal{X})$ is the residual of the receiver clock factors. $\sigma_{r,p}^s$ is the pseudorange variance. $D_{t,k}$ is a covariance matrix associate with the residual of receiver clock factors.

The anchor point coordinate and the receiver clock biases associate with each GNSS epoch are refined through the optimization of the above problem. After this step, the anchor point, origin of the ENU frame, is set to the origin of the local world frame. Finally, the initialization phase of the entire estimator is finished and all necessary initial quantities have been generated to boot the system up.

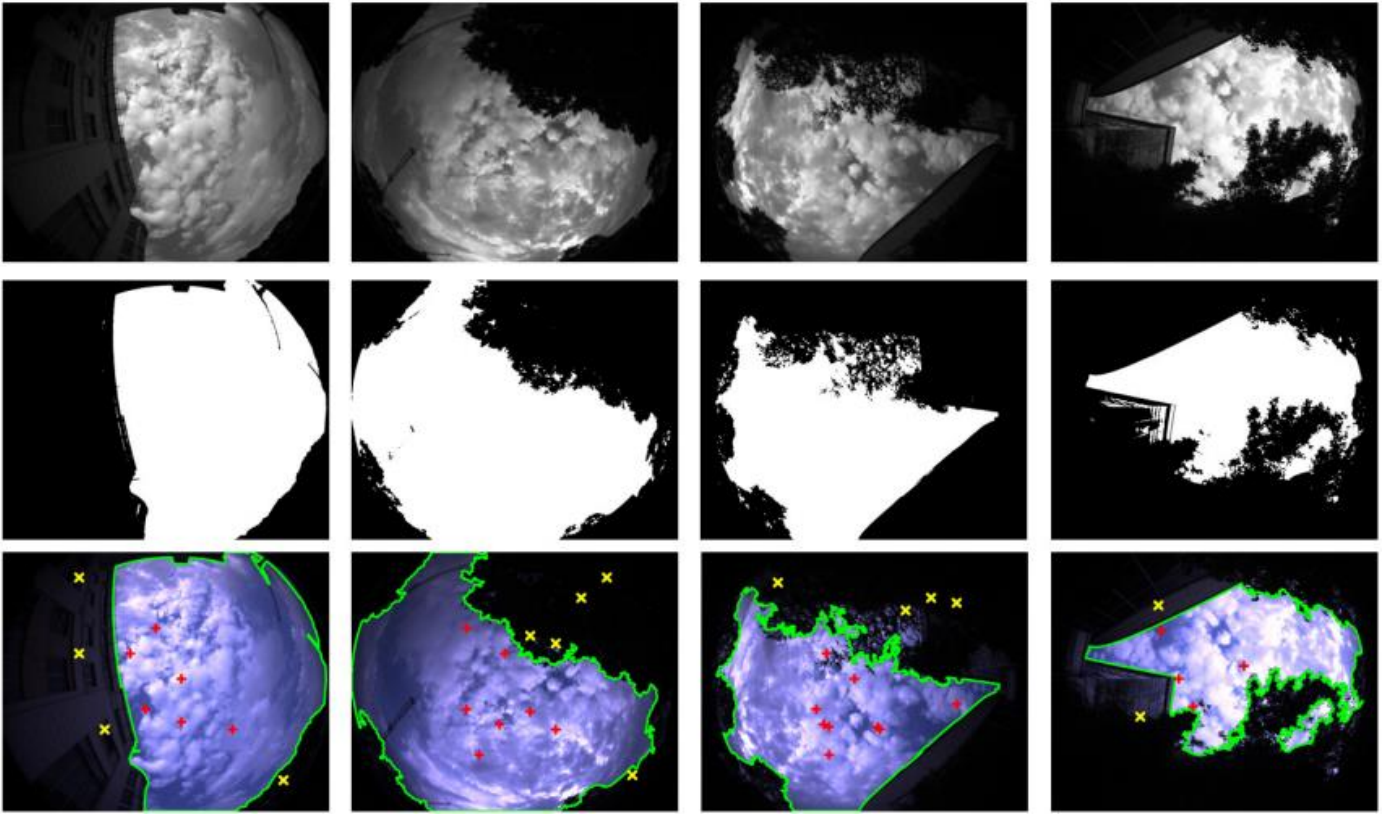

Figure 2. Results of Sky Segmentation and Back Projection. Images in the first row are grey-scale images after blurring. Images in the second row are after thresholding. Images in the third row are sky segmentation results, where green borders represent the border of sky regions and non-sky regions, red points represent LOS satellites, and yellow points represent NLOS satellites.

*3.4. NLOS Mitigation*

GNSS positioning accuracy is greatly affected by unhealthy satellite measurement. Therefore, improving satellite signal quality is critical. In GVINS, multi-path effect is mitigated by excluding the low elevation satellites. Furthermore, Satellite filtration is also achieved through weight setting of factor graph optimization. More specifically, the pseudo-range noise $\varepsilon_r^s$ is assumed to be zero-mean Gaussian distributed such as $\varepsilon_r^s \sim N(0, \sigma_{r,p}^s)$, where the variance $\sigma_{r,p}^s$ is modelled as:

$$\sigma_{r,p}^s = \frac{n_{si} \times n_p}{\sin^2 \theta_{el}} \quad (5)$$

where $n_{si}$ is the broadcast satellite space accuracy index, and $n_p$ is the pseudo-range measurement noise index obtained by the receiver. $\theta_{el}$ is the satellite elevation angle from the receiver.

Similarly, the Doppler measurement noise $\eta_{r,d}^s$ is assumed to be Gaussian distributed and the corresponding

variance is modelled as:

$$\sigma_{r,d}^s = \frac{n_{si} \times n_d}{\sin^2 \theta_{el}} \quad (6)$$

where $n_d$ is the measurement noise index reported by the receiver.

As indicated in above equations, satellites with higher noise variance will be assigned a lower weight in the subsequent optimization procedure. However, NLOS effect has not been considered in the GNSS calculation phase in (Cao et al. 2021), which might contribute to a larger localization error. To address this issue, we propose a sky segmentation based NLOS exclusion strategy. We firstly perform a back-projection of satellite coordinate to sky-pointing image frame and then we segment the sky-pointing image. Below are the implementing details of our strategy.

1) Back projection: We obtain 3D coordinates of satellites $p_s^e$ from a Ublox F9P receiver in the ECEF frame and simultaneously capture the sky image via a fish-eye camera. In the initialization procedure, the visual-inertial odometry has been aligned with GNSS trajectory via an anchor point. Then the receiver's ECEF coordinate can be mapped to the corresponding coordinate in the local world frame via this anchor point, at which the ENU frame is built. Below are detailed mathematical procedures:

We set the anchor point to the origin of the local world frame, which coincides with the origin of the ENU frame. The rotation from the ECEF frame to the ENU frame is

$$R_e^n = \begin{bmatrix} -\sin \lambda & \cos \lambda & 0 \\ -\cos \lambda \sin \phi/2 & -\sin \phi \sin \lambda/2 & \cos \phi \\ \cos \lambda \cos \phi/2 & \cos \phi \sin \lambda/2 & \sin \phi \end{bmatrix} \quad (7)$$

where $\phi$ and $\lambda$ is respectively the latitude and longitude of the anchor point in geographic coordinate system. Given the ENU coordinate of the anchor point, the satellite coordinate in the ENU frame can be expressed as:

$$p_s^n = R_e^n p_s^e + p_{anc}^n \quad (8)$$

The 1-DOF rotation between the ENU and the local world frame $R_w^n$ given by the yaw offset $\psi$. The relationship between the local world frame and the ENU frame is given by:

$$R_n^w = R_w^{n-1} \quad (9)$$

Then the satellite coordinate in the local world frame can be expressed as:

$$p_s^w = R_n^w p_s^n + p_{anc}^w \quad (10)$$

As we have previously finished the calibration, knowing the extrinsic parameters $R_i^{sky}$ between forward-looking camera and IMU and the extrinsic parameters $R_i^w$ between sky-pointing camera and IMU, the relationship between sky-pointing camera frame and local world frame can be expressed as:

$$R_{sky}^w = R_i^w R_{sky}^i \quad (11)$$

The satellite coordinates in the sky-pointing camera frame can be expressed as:

$$p_s^{sky} = R_w^{sky} p_s^w \quad (12)$$

Furthermore, the receiver's antenna and the sky-pointing fish-eye camera are mounted with a distance of merely several centimeters on our robot. Therefore, the offset between them can be omitted. This is, we do not distinguish between $p_r^w$ and $p_s^w$.

So far we have obtained the 3D coordinates of satellites in the sky-pointing fish-eye frame. Considering the large distortion caused by the fish-eye lens, we need to implement the distortion process before we project the 3D coordinates to the image plane of the fish-eye camera. Assuming the 3D point of satellite with ID $i$ in the sky-pointing camera frame is $p_s^{sky}$, then the pinhole projection coordinates satellite $i$ is [a; b], where

$$\begin{aligned} a &= x/z \text{ and } b = y/z \\ r^2 &= a^2 + b^2 \\ \theta &= \text{atan}(r) \end{aligned} \quad (13)$$

We apply Kannala Brandt fish-eye model (Kannala et al. 2006) for distortion:

$$\theta_d = \theta(1 + k_1\theta^2 + k_2\theta^4 + k_3\theta^6 + k_4\theta^8) \quad (14)$$

The distorted point coordinates are $[x', y']$ where

$$\begin{aligned} x' &= (\theta_d/r)a \\ y' &= (\theta_d/r)b \end{aligned} \quad (15)$$

Finally, convert distorted point coordinates into pixel coordinates with following equations:

$$\begin{aligned} u &= f_x(x' + \alpha y') + c_x \\ v &= f_y y' + c_y \end{aligned} \quad (16)$$

2)Sky Segmentation: Although some learning-based methods can achieve high accuracy, they typically require a lot of data with ground truth labels to train on. Also, they involve higher processing complexity and power costs, which are not suitable for real-time SLAM systems which are strict on the calculation efficiency. Therefore, we attempt to focus on conventional methods which are more lightweight. We adopt the Otsu algorithm, whose effectiveness will be verified in subsequent experiments.

Firstly, we match each GNSS measurement with a corresponding sky-pointing fish-eye image according to their timestamps. Secondly, we convert the acquired color image to a grey-scale image. Then we blur the image using the mean filtering function to suppress and prevent interference and noise points. After that, we go into the critical part: Calculating an appropriate threshold for the image.

The threshold method segments an image into similar regions according to predefined criteria. The selection of the threshold is of vital significance to the segmentation results. Mathematically, it can be expressed as:

$$R(i,j) = \begin{cases} 255 & \text{if } I(i,j) > \text{threshold} \\ 0 & \text{else} \end{cases} \quad (17)$$

where $I(i,j)$ and $R(i,j)$ represent the pixel value in the $i$-th row and $j$-th column of the image before and after thresholding respectively. Here we introduce a classic commonly used global threshold method.

---

**Algorithm 1** The Otsu Method
---
**Input:** An image to be segmented
**Parameter:** Weights $\omega_0$ and $\omega_1$ are the probabilities of the two classes separated by a threshold $t$, and $\sigma_0^2$ and $\sigma_0^1$ are variances of these two classes.
1: Compute histogram and probabilities of each intensity level.
2: Set up initial $\omega_i(0)$ and $\mu_i(0)$.
3: Step through all possible thresholds t=1,...,maximum. Update $\omega_i$ and $\mu_i$ and compute $\sigma_b^2(t)$
4: Output the threshold corresponding to the maximum $\sigma_b^1(t)$

---

Otsu algorithm is a global threshold algorithm based on Fisher's criterion (Malina et al. 1981) – i.e. selecting a comprehensive discriminant variable or projection direction so that all kinds of points are concentrated as far as possible, and classes and classes are separated as far as possible to minimize the intra-class variance and maximize the inter-class variance. The pipeline of the Otsu method is shown in Algorithm 1. Assuming the threshold $t$ divides all pixels of the image into two types $C1$ (less than $t$) and $C2$ (greater than $t$). Then the mean values of these two types of pixels are respectively $\mu_0(t)$ and $\mu_1(t)$, and the mean value of the whole image is $\mu_T$. They can be respectively formulated as:

$$\mu_0(t) = \frac{\sum_{i=0}^{t-1} ip(i)}{\omega_0(t)}$$
$$\mu_1(t) = \frac{\sum_{i=t}^{L-1} ip(i)}{\omega_1(t)} \quad (18)$$
$$\mu_T = \sum_{i=0}^{L-1} ip(i)$$

Meanwhile, the class probability of pixel being classified into $C1$ and $C2$ is $\omega_0(t)$ and $\omega_1(t)$, which can be respectively calculated by:

$$\omega_0(t) = \sum_{i=0}^{t-1} p(i)$$
$$\omega_1(t) = \sum_{i=t}^{L-1} p(i) \quad (19)$$

where $p(i)$ represents the probability of pixel value $i$ and $L$ represents the maximum pixel value. We can obtain the following relations:

$$\begin{aligned} \omega_0\mu_0 + \omega_1\mu_1 &= \mu_T \\ \omega_0 + \omega_1 &= 1 \end{aligned} \quad (20)$$

The class probabilities and class means can be computed iteratively based on above equations. And then, the algorithm exhaustively searches for the threshold that minimizes the intra-class variance, which is defined as a weighted sum of variances of the two classes:

$$\sigma_w^2(t) = \omega_0(t)\sigma_0^2(t) + \omega_1(t)\sigma_1^2(t) \quad (21)$$

where $\sigma_0^2$ and $\sigma_1^2$ are variances of these two classes. For two classes, minimizing the intra-class variance is equivalent to maximizing inter-class variance (Otsu et al. 1979):

$$\begin{aligned} \sigma_b^2(t) &= \sigma^2 - \sigma_w^2(t) = \omega_0(t)(\mu_0 - \mu_T)^2 + \omega_1(t)(\mu_1 - \mu_T)^2 \\ &= \omega_0(t)\omega_1(t)[\mu_0(t) - \mu_1(t)]^2 \end{aligned} \quad (22)$$

Having known the satellite coordinates and sky regions, we can treat satellites in the sky regions as LOS signals and those in non-sky regions as NLOS signals.

## 4. Experiments and Evaluation

### 4.1. Acquisition Platform

We construct a ground robot as shown in Figure 3. A sky-pointing fish-eye camera is mounted on the top layer to capture sky images in a large FOV (Field Of View). A VI-sensor (Visual-Inertial Sensor) records forward-looking images and high-rate IMU information. A GNSS receiver collects GNSS raw measurements and a high-accuracy GNSS-IMU device with RTK signals to obtain ground truth trajectory. The GNSS-IMU device obtains a high localization accuracy even with RTK signals lost for a few seconds.

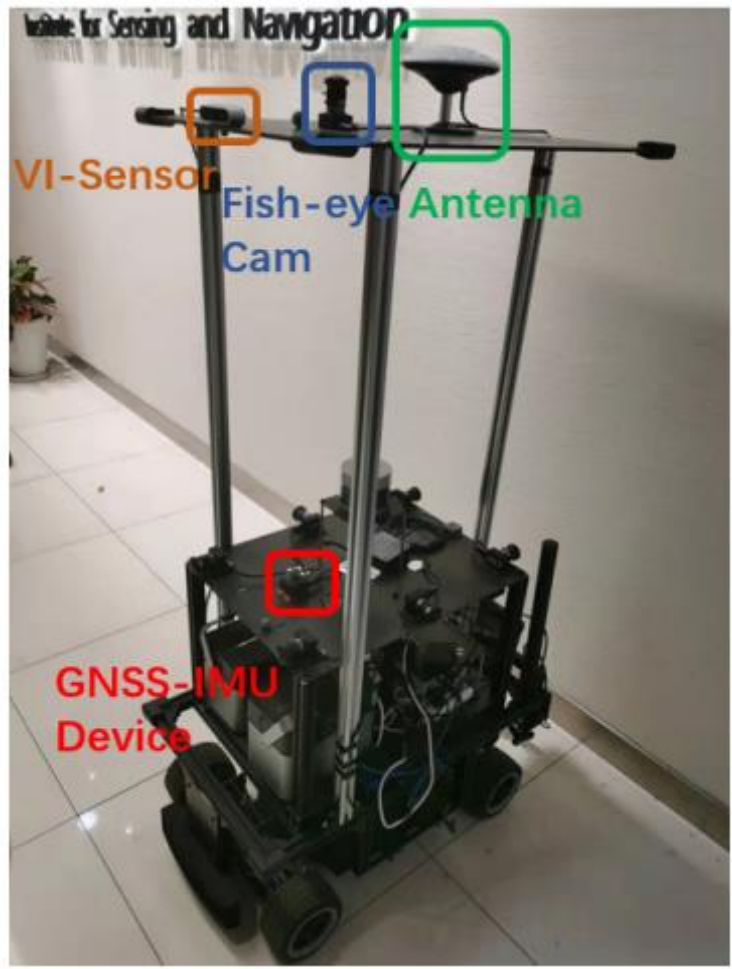

Figure 3. Our ground robot for experimental data collection. The platform is
equipped with a forward-looking VI-sensor, a sky-pointing fish-eye camera
and a high-accuracy GNSS-IMU device.

The detailed parameters of our sensors are given in Table 1. We use the Kalibr toolbox (Furgale et al. 2013) to calibrate the extrinsic parameters of cameras and IMU, MATLAB camera calibration toolbox (Pinhole Model [Strob et al. 2011]) to calibrate pinhole camera, and Kannala Brandt Model (Kannala et al. 2006) to calibrate the fish-eye camera.

Table 1. Specifics of sensors and tracking devices.

| Device | Type | Spec. | Freq.(Hz) |
|---|---|---|---|
| GNSS Reciver | Ublox Zed-F9P | GPS/BeiDou/GLONASS/GALILEO | 1 |
| RGB Camera | FLIR Pointgrey CM3-U3-13Y3C | 1280*1024, 190 H-FOV, 190 V-FOV | 30 |
| VI-Sensor | Realsense D435I | RGB: 640*480, 69 H-FOV, 42.5V-FOV | 30 |
|  |  | IMU: 6-axis | 200 |
| RTK/INS | Xsens Mti 680G | RTK: localization accuracy 2cm | 100 |
|  |  | INS: 9-axis | 100 |

Note: In this table, the bold indicates the best result of the objective index in these algorithms.

*4.2. Sky Segmentation Test*

The vision-based NLOS exclusion method applied in this research relies heavily on the accuracy of the segmentation algorithm. If LOS measurements are filtered by mistake, it might have a negative impact on the location results. To test the capability of segmenting sky regions, we capture 500 sky images in different areas and label the sky regions on these images with Labelme Toolbox (Wkentaro et al. 2019). For threshold methods, we try both global threshold Method (Otsu [Otsu et al. 1979]) and local threshold Method (Ahmad et al. 1999). In Otsu method, we fine-tune the threshold to generate the best accuracy.

For clustering methods, we test the supervised classifier SVM (Cherkassky et al. 2004) and the unsupervised classifier K-means (Krishna et al. 1999). Among them, SVM and GrabCut need previous information about foreground and background regions before segmenting each image. We do not manually select these regions one by one for each image but select a fixed image range for them. Because sky regions do not remain in the same or close position in different images, these two methods obtain a bad segmentation result.

These algorithms are tested on our dataset with 500 sky-pointing images which were captured on campus under buildings, trees or hardly with occlusions in the sky and the performance is shown in Table 2. The classification rate is measured by IOU (Intersection Over Union) (Rezetofighi et al. 2019) of sky-region areas. Results show Otsu algorithm has less time consumption and a higher classification rate than other algorithms in the test. We analyze that the sky regions are featured with blue and white pixel blocks compared with non-sky regions. The global threshold algorithm can distinguish the pixel blocks with special pixel information well. Unlike global threshold Methods, the local threshold Method chooses different threshold values for every pixel in the image based on an analysis of its neighboring pixels, which is not suitable for obtaining a global threshold to segment the sky regions and the non-sky regions in the whole image. As a result, we choose the Otsu algorithm for our system.

Table 2. Performance of different image segmentation sky dataset

| Method | OTSU | Local | GrabCut | SVM | K-means |
|---|---|---|---|---|---|
| Time(s) | **0.0105** | 0.0081 | 0.0746 | 1.9204 | **0.5225** |
| IOU(%) | **96.5453** | 68.6905 | 51.6352 | 60.5884 | 89.3684 |

Table 3. ATE(meter) RMSE of SLAM systems on sample sequences

| Method | Street01 | Street02 | Street03 | Bridge01 |
|---|---|---|---|---|
| VINS-Mono | 15.44 | 24.72 | 7.86 | 17.14 |
| GVINS | 8.23 | 25.06 | 7.85 | **3.84** |
| Sky-GVINS(L) | 29.28 | 27.93 | 7.89 | 11.74 |
| Sky-GVINS(O) | **7.91** | **4.27** | **7.82** | **3.84** |

*4.3. Real-world Experiment*

To fully evaluate our proposed system, we operated our robot cross on the Shanghai Jiao Tong University campus and simultaneously recorded experimental trajectories in various scenarios. In dense urban environments, we collected Sequences *Street01* and *Street02* with many occlusions including trees and buildings. For open-sky tests, we recorded Sequence *Street03* on a wide road with hardly any occlusions in the sky, while recording Sequence *Bridge01* on a bridge back and forth. At the beginning of all the sequences, the ground robot traveled some distance in a straight line over an open area to ensure the successful initialization of GVINS and Sky-GVINS. A demo video of Sky-GVINS in sequence *Street02* is displayed on the website https://www.youtube.com/watch?v=XGGV9fB7raA.

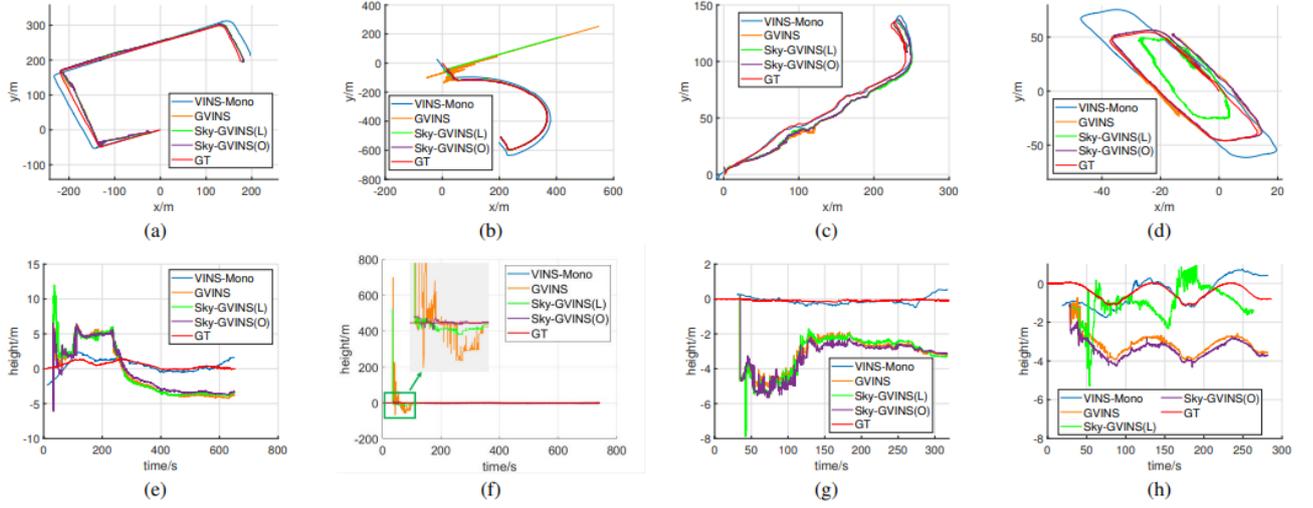

Figure 4. Estimated and ground-truth (GT) trajectories of 7 sample sequences are visualized in ENU (East-North-Up) coordinate system. (a)~(d): x-y graph of *Street01*, *Street01*, *Street03* and *Bridge01* respectively. (e)~(h): time-height graph of *Street01*, *Street01*, *Street03* and *Bridge01* respectively.

### 4.4. Evaluation and Analysis

We run VINS-Mono (Qin et al. 2018), GVINS (Cao et al. 2021) and Sky-GVINS with different segmentation algorithms on our self-collected datasets. To test the impact of sky segmentation accuracy on the final result, we test the K-means algorithm but it can not run in real-time due to its high time consumption. Therefore, we try local threshold method instead. For simplicity, in the following paragraphs, we denote Sky-GVINS with the Otsu segmentation and Sky-GVINS with the local threshold segmentation as Sky-GVINS(O) and Sky-GVINS(L) respectively.

The Root Mean Squared Error (RMSE) of Absolute Trajectory Error (ATE) (Sturm et al. 2012) is used for the evaluation metric. We visualize all the estimated trajectories and ground truth trajectories with different colors as shown in Figure 4. Since GVINS (Cao et al. 2021) and Sky-GVINS can directly output estimation results in the ECEF frame, we do not apply any alignment to their trajectories. For VINS-Mono (Qin et al. 2018) which merely outputs its trajectory in the local frame, a 6-DOF alignment is performed with the EVO tool (Michael et al. 2018) to align the estimated trajectories with the ground truth to obtain the ATE Error.

In all the four tested sequences, Sky-GVINS(O) effectively filters NLOS signals and outperforms all the other tested systems. More specifically, in urban canyons like Sequence *Street01*, GNSS based-methods have better positioning results in plane, but there is a certain drift in height. While VINS-Mono remains relatively stable in height and has a large error in plane positioning in this case. In Sequence *Street02*, GVINS and Sky-GVINS(L) drift severely after initialization due to inaccurate satellite signals and produce a bad ATE result, as shown in Figure 4 (b) and (f). In the open-sky Sequence Street03, all the tested algorithms achieve close accuracy, with Sky-GVINS(O) narrowly outperforming the others. This indicates that in this sequence, most of the received signals are form LOS satellites and the VIO contributes a lot to the final localization result. Even so, Sky-GVINS(O) still successfully eliminates a small number of NLOS signals, resulting in a minor improvement in positioning accuracy. In Sequence Bridge01, Sky-GVINS(O) obtains an equally good result as GVINS. From Figure 4 (h), we can observe that there is a fixed height offset of about 2 meters between GVINS and Sky-GVINS trajectories and the ground truth since the starting point of this sequence, which accounts for a large part of the total ATE error. This error comes from the alignment of VIO and GNSS during the initialization phase, which demonstrates the importance of initialization accuracy.

Furthermore, we notice that even with NLOS mitigation at high accuracy of sky segmentation, the height drift of Sky-GVINS(O) has not been completely eliminated as shown in Figure 4 (e) and (g). Comparing VINS-Mono's trajectory in Figure 4, we know that VIO trajectories usually remain relatively stable at height. Therefore, the deviation in height of Sky-GVINS is largely due to the error in GNSS calculation. We analyze there might be three reasons: First, inaccurate pseudo-range measurement in dense urban environments will degrade the precision of pseudo-range positioning, which may lead to the deviation of the satellite coordinates obtained by the Ublox receiver and the satellite position back-projected on the image. Second, there exists a minor time offset between GNSS measurement and the corresponding sky-pointing image, during which the ground robot may have moved a certain distance. This gap can be filled by a hardware-synchronized sensor suite of a camera and a GNSS receiver. Third, due to the previous incorrect satellite exclusion, there is a deviation in the position estimation, which will affect the position of the satellite back-projected on the image, and result in the incorrect satellite exclusion again, forming a vicious circle. The third possibility in particular emphasizes the importance of the initialization accuracy of the anchor point's position.

## 5. Conclusions

We proposed Sky-GVINS: a robust and reliable localization method with low-cost sensors. Extensive experiments show our system can achieve robust localization in open-sky areas and dense urban environments. Furthermore, we implement a lightweight sky-segmentation and NLOS mitigation method without dependency on previously known 3D models or the selection of parameters. This method can be easily extended to other GNSS-based methods with one requirement: Orientation can be determined by IMU pre-integration or dual antenna to obtain the 6-DOF relationship of the GNSS receiver and camera. Future work will consider tightly coupling the sky-pointing image to the system because most objects captured at this angle are static instead of moving, which is a helpful feature for navigation in highly dynamic environments.

**Funding**

This work was supported by National Key R&D Plan of China (2022YFB3903800) and NSFC(62073214).

**Data Availability Statement**

The data support the findings of this study are openly available in https://github.com/sjtuyinjie/Sky-GVINS-Dataset.

**Disclosure Statement**

There are no financial conflicts of interest to disclose.

**Notes on contributors**

*Jie Yin* received the B.S. degree in the electrical engineering from Shanghai Jiao Tong University, Shanghai, China, in 2021. And he is currently pursuing the master degree with Shanghai Jiao Tong University, Shanghai, China. His current research interests include Computer Vision, Robotics, Multi-sensor Fusion, and Simultaneous Localization and Mapping (SLAM).

*Tao Li* received the B.S. degree in navigation engineering from Wuhan University , Wuhan, China, in 2018. He is currently pursuing the Ph.D. degree with Shanghai Jiao T ong University ,Shanghai, China. His current research interests include Visual-SLAM, LiDAR-SLAM, Global Navigation satellite systems, inertial navigation systems, and information fusion.

*Hao Yin* graduated from Shanghai Jiao Tong University with a Bachelor's degree in Industrial Engineering in 2021. He is currently studying for a master's degree at Tsinghua University in China. His current research interests include combinatorial optimization, robotics, etc.

*Wenxian Yu* is Wenxian Yu(Senior Member, IEEE) received the B.S., M.S., and Ph.D. degrees from the National University of Defense Technology, Changsha, China, in 1985, 1988, and 1993,respectively . From 1996 to 2008, he was a Professor with the College of Electronic Scienceand Engineering, National University of Defense Technology , where he was also the Deputy Head of the College and the Assistant Director of the National Key Laboratory of Automatic Target Recognition. From 2009 to 2011, he was the Executive Dean of the School of Electronic, Information, and Electrical Engineering, Shanghai Jiao Tong University , Shanghai, China. He is currently a Y angtze River Scholar Distinguished Professor and the Head of the research part in the School of Electronic, Information, and Electrical Engineering, Shanghai Jiao Tong University. His research interests include remote sensing information processing, automatic target recognition, and multi-sensor data fusion.

*Danping Zou* received his Ph.D. in Computer Application Technology from the School of Computer Science at Fudan University in 2010 and was a postdoctoral researcher at the National University of Singapore from 2010 to 2013. In 2013, he joined the Institute of Perception and Navigation, Shanghai Jiao Tong University. He is currently an associate professor at Shanghai Jiao Tong University. His research interests include real-time computer 3D vision, Simultaneous Localization and Map Building, and autonomous navigation for unmanned systems.